\documentclass[sigconf]{acmart}
\makeatletter
\def\@ACM@checkaffil{
    \if@ACM@instpresent\else
    \ClassWarningNoLine{\@classname}{No institution present for an affiliation}%
    \fi
    \if@ACM@citypresent\else
    \ClassWarningNoLine{\@classname}{No city present for an affiliation}%
    \fi
    \if@ACM@countrypresent\else
        \ClassWarningNoLine{\@classname}{No country present for an affiliation}%
    \fi
}
\makeatother

\usepackage{multirow}
\usepackage[font=footnotesize]{subfig}

\AtBeginDocument{%
  \providecommand\BibTeX{{%
    \normalfont B\kern-0.5em{\scshape i\kern-0.25em b}\kern-0.8em\TeX}}}

\renewcommand\footnotetextcopyrightpermission[1]{} 

\acmSubmissionID{4264}

\begin{document}
\pagestyle{plain}
\title{Compressed Deepfake Video Detection Based on 3D Spatiotemporal Trajectories}

\author{Zongmei Chen}
\email{chenzongmei@hnu.edu.cn}
\affiliation{%
  \institution{Hunan University, China}
}

\author{Xin Liao}
\authornote{Corresponding author.}
\email{xinliao@hnu.edu.cn}
\affiliation{%
	\institution{Hunan University, China}
}

\author{Xiaoshuai Wu}
\email{shinewu@hnu.edu.cn}
\affiliation{%
  \institution{Hunan University, China}
}

\author{Yanxiang Chen}
\email{chenyx@hfut.edu.cn}
\affiliation{%
  \institution{Hefei University of Technology, China}
}



\begin{abstract}
  The misuse of Deepfake technology by malicious actors poses a potential threat to nations, societies, and individuals. However, existing methods for detecting Deepfakes primarily focus on uncompressed videos, such as noise characteristics, local textures, or frequency statistics. When applied to compressed videos, these methods experience a decrease in detection performance and are less suitable for real-world scenarios. In this paper, we propose a Deepfake video detection method based on 3D spatiotemporal trajectories. Specifically, we utilize a robust 3D model to construct spatiotemporal motion features, integrating feature details from both 2D and 3D frames to mitigate the influence of large head rotation angles or insufficient lighting within frames. Furthermore, we separate facial expressions from head movements and design a sequential analysis method based on phase space motion trajectories to explore the feature differences between genuine and fake faces in Deepfake videos. We conduct extensive experiments to validate the performance of our proposed method on several compressed Deepfake benchmarks. The robustness of the well-designed features is verified by calculating the consistent distribution of facial landmarks before and after video compression. Our method yields satisfactory results and showcases its potential for practical applications.
\end{abstract}



\keywords{Compressed Deepfake videos, facial expressions, head movements, 3D model}


\maketitle
\begin{figure}[h]
	\centering
	\subfloat[Real-HD]{\includegraphics[width=0.46\columnwidth]{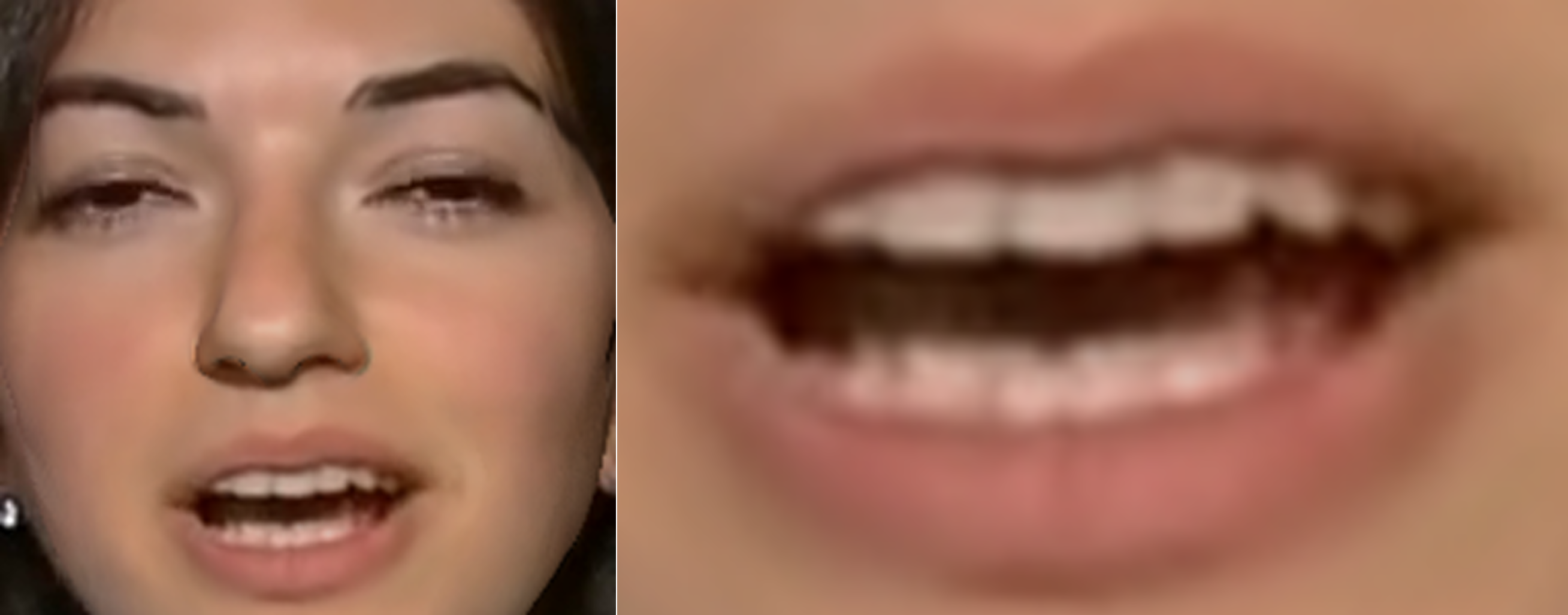}}\hspace{3mm}
	\subfloat[Fake-HD]{\includegraphics[width=0.47\columnwidth]{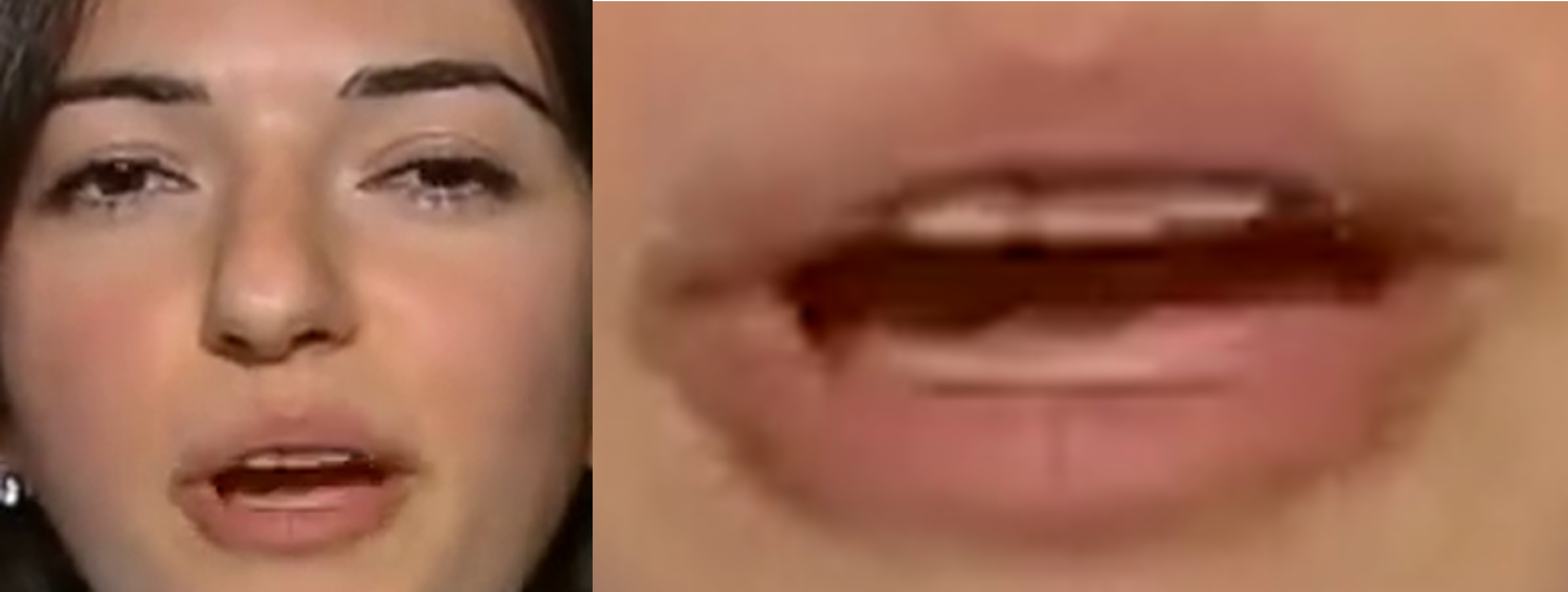}}
 \\
 \subfloat[Real-Compressed]{\includegraphics[width=0.46\columnwidth]{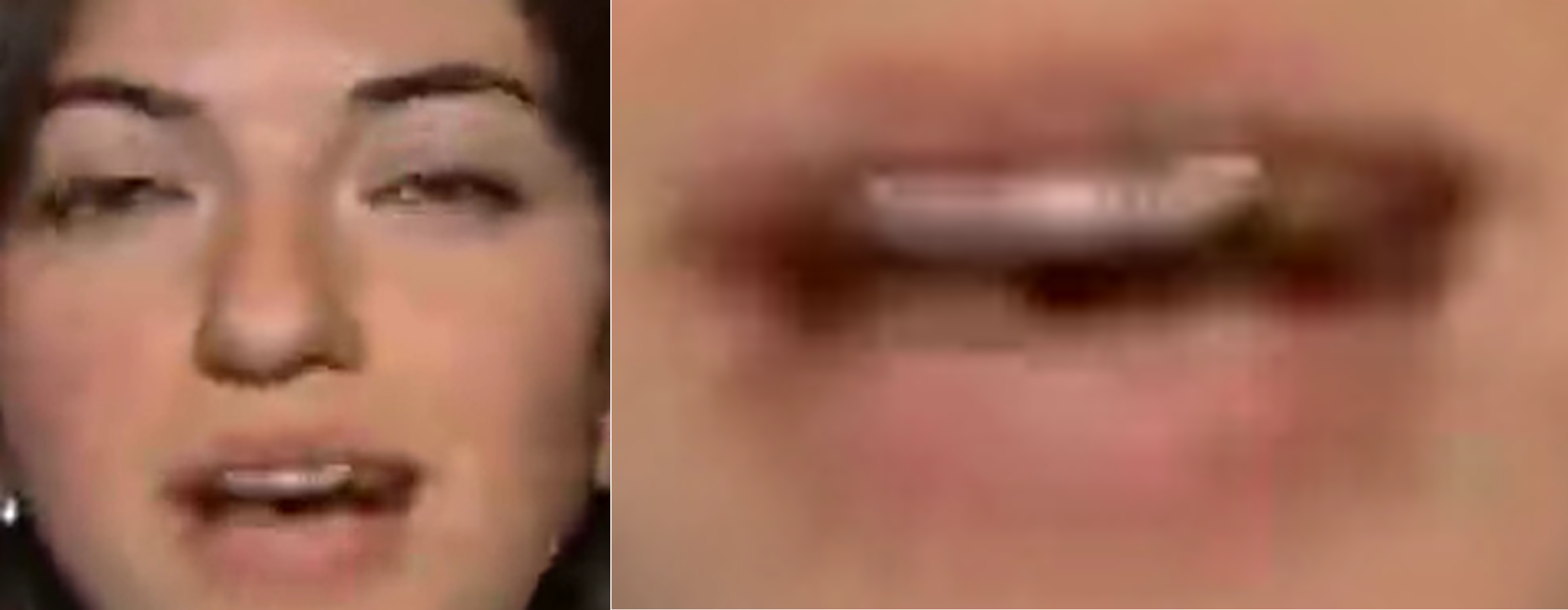}}\hspace{3mm}
	\subfloat[Fake-Compressed]{\includegraphics[width=0.47\columnwidth]{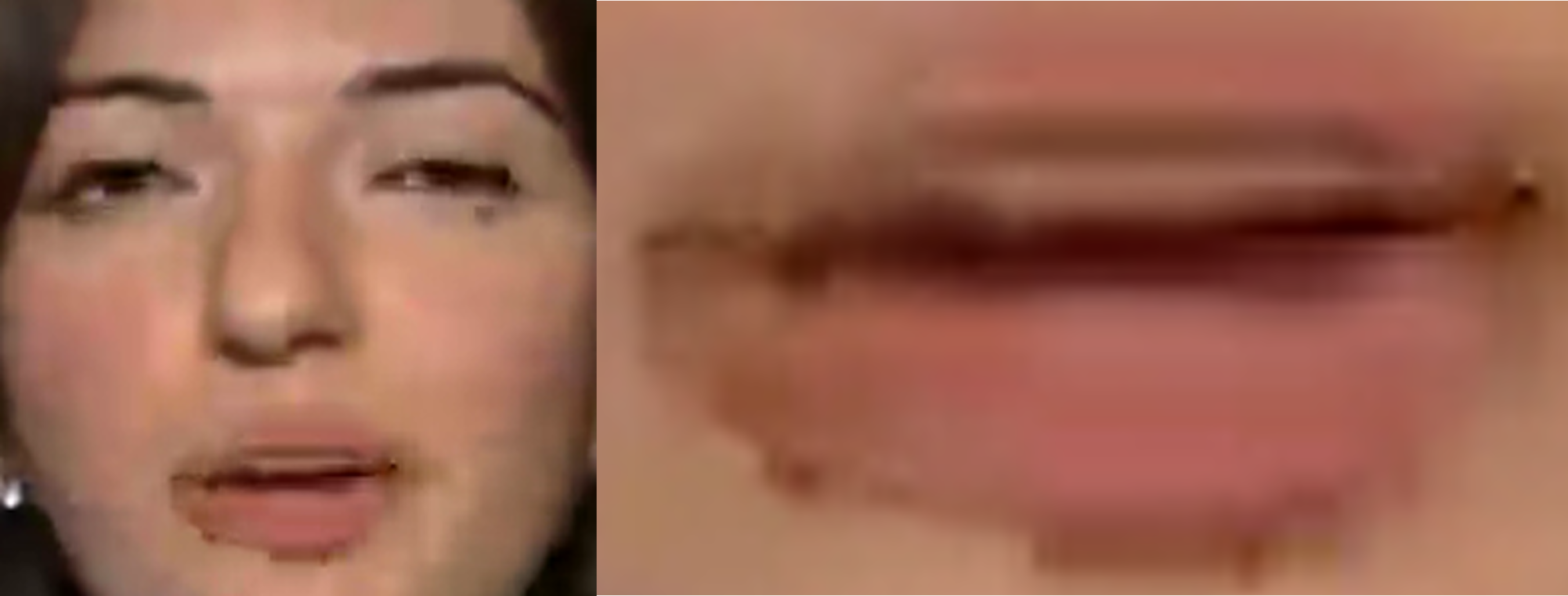}}
	\caption{Visualization of real and fake videos in compressed and uncompressed states. The videos are from FaceForensics++. HD stands for high definition. Comparing the first and second columns, the lip borders and teeth of the fake video become blurred and tampering artifacts are present. Comparing the first and second rows, the lips and teeth lose their obvious shape in the compressed video, and compression artifacts appear. When compression artifacts and tampering artifacts coexist, the teeth are no longer visible and the shape of the lips has changed. These challenges lead to the low accuracy observed in current methods for detecting compressed Deepfake videos.}
 \label{fig:vis}
\end{figure}
\section{Introduction}
Deepfakes \cite{Deepfake} is a compound word that is a combination of Deep Learning and Fake in the field of artificial intelligence (AI). It is often used to describe fake media generated using deep learning and other technologies. Deepfake technology can not only replace the identity of the target person in images and videos, but also allow the target person to make corresponding expressions based on the driving video or specified audio. In addition, Deepfake technology can also edit the facial attributes of the target person, and even generate faces that do not exist in real life. In recent years, deep learning algorithms have been
continuously iterated, artificial intelligence-generated content
has flourished, and high-quality images and videos forged by
AI have reached a level that is indistinguishable to the human eye. Deepfake technology has certain positive application value. This technology can promote the emerging development of the entertainment and cultural exchange industry
and has strong entertainment and communication properties.
However, some criminals use these technologies to commit
telecom fraud, create fake news, slander celebrities, publish
false statements, destroy identity verification, sell pornography, etc., posing serious threats to individuals, society, and the country. Therefore, it is particularly critical to carry out research on Deepfake detection technology.

Current Deepfake video detection methods are mainly categorized into active defense and passive detection. Active defense methods prevent the generation of Deepfake videos by adding signal interference. However, the execution conditions of these methods are harsh, and the data source is not controllable. On the other hand, passive detection methods can be divided into learning from forged samples \cite{Two-stream,Xception,Efficientnet,Two-branch}, learning without forged samples \cite{Facex-ray,52zhao2021learning,r8}, task migration \cite{LipForensics,r10,r11}, and generate data-driven \cite{r12,r13}. These methods explore identifiable facial features from various perspectives, including spatial, frequency, multimodal domains, etc. However, due to the complexity of propagation scenarios and the presence of multiple adversarial factors, these methods show poor robustness and generalization ability in real forgery scenarios.

Video compression on social networks is a common phenomenon. When users upload videos to social media platforms (such as Facebook, Instagram, Twitter, etc.), these platforms usually compress the video to reduce the file size and
speed up the upload and playback. This compression usually uses different compression algorithms and
parameters to balance video quality and file size. As depicted in Fig. ~\ref{fig:vis}, compression may cause the loss of video details, reduce the resolution,
frame rate and visual quality of the video, causing the overlap of compression artifacts and tampering artifacts \cite{zhang2022}, thereby reducing the performance of existing Deepfake video detection methods.

To address compressed Deepfake videos in real-world scenarios, this paper proposes a detection method based on 3D spatiotemporal trajectories to enhance the detection performance and robustness of compressed Deepfake video detection. In detail, by studying facial motion, constructing motion features in the temporal domain and spatial domain, and performing time series analysis on phase space motion trajectories to realize the authenticity determination of Deepfake videos. The main contributions of this paper include:
\begin{enumerate}
\item{We propose a spatiotemporal feature construction method based on the robust 3D model, which directly locates and tracks facial and head landmarks in the video, and combines the spatial dynamics and temporal characteristics of facial action units to construct the features. To the best of our knowledge, we are the first to migrate 3D models to facial landmark localization and tracking in Deepfake video detection.}
\item{A temporal feature analysis method based on phase space motion trajectories is proposed to model the facial change pattern between the first frame and each subsequent frame within a continuous period of time. It explores the temporal changes of facial coordinates in 3D space, and analyze the overall and global characteristics of the video.}
\item{We conduct extensive experiments to verify the detection performance on compressed Deepfake videos and the results demonstrate the propsed achieves promising performance compared to the state-of-the-art methods. In addition, our method is able to avoid the effects of large head rotation angle or low illumination. It better resolves the task of detecting Deepfake videos in real scenes.}
\end{enumerate}
\section{Related Work}
\subsection{Deepfake generation techniques}
Building on Generative Adversarial Networks (GANs) \cite{goodfellow2014generative}, Deepfake techniques have progressed. It showing advantages in image synthesis, especially in facial replacement and background integration. FaceSwap \cite{FaceSwap} represents a typical computer graphics-based Deepfake synthesis algorithm. This algorithm utilizes techniques such as 3D face reconstruction models and affine transformations. However, it requires a high degree of similarity between the source and target images and may leave noticeable fake traces. Another computer graphics-based method, proposed by Thies et al. \cite{face2face}, involves real-time facial capture and reproduction. Nevertheless, alignment issues arise due to incomplete consistency between the head postures of the target subject and the source subject. In contrast, synthesis methods based on deep learning techniques are more prevalent. Deepfakes \cite{Deepfakes} introduces deep learning techniques for the first time to the task of face synthesis. It employs a shared encoder to encode and decode the source and target images to generate fake face images. FaceShifter \cite{li2020advancing} presents a novel two-stage framework for high-fidelity and occlusion-aware face swapping. However, these models generate faces with low resolution and lack texture details. To address this issue, researchers have proposed various improvements, such as local-global dual-branch networks \cite{xu2022region}, the X2Face model \cite{wiles2018x2face}, occlusion-aware networks \cite{siarohin2019first}, and the Wav2Lip model \cite{prajwal2020lip}. Furthermore, progressive training networks like ProGAN \cite{ProGAN} and generator architectures like StyleGAN \cite{karras2019style} have provided crucial insights and methods for the development of Deepfake technology. The emergence of these methods has led to significant progress in the field of visual synthesis through Deepfake technology.
 \subsection{Deepfake detetion techniques}
To our knowledge, researchers have proposed four types of detection technologies to determine whether the video is Deepfake. 

\textbf{Learning from forged samples.} Two-stream \cite{Two-stream} leverages steganalysis features from traditional image forensics, while XceptionNet \cite{Xception} and EfficientNet \cite{Efficientnet} extract spatial features from frames. The Two-branch \cite{Two-branch} structure constructs a dual-branch network architecture to achieve multi-domain information fusion. The core characteristic of such methods is to utilize paired genuine and fake data as the driving force for training. The learning process of the classification model requires the involvement of synthesized facial samples. However, it exhibits strong data dependency, weak generalization, and significant impacts on model performance from unknown tampering types and compression. 

\textbf{Learning without forged samples.} Face X-ray \cite{Facex-ray} detects traces of fusion operations in forgery methods. PCL \cite{52zhao2021learning} measures the consistency between the source features of face images. Guillaro et al. \cite{r8} use a fusion architecture based on Transformer to extract high and low-level traces. This type of model training process does not require the use of synthesized negative samples of faces as data drivers. Instead, it captures certain characteristics of the facial information carrier or exploits inherent process vulnerabilities in Deepfakes to achieve detection and authentication. Because it does not rely on paired genuine and fake facial data, it exhibits strong transfer detection capabilities across different forgery algorithms, and its cross-dataset detection performance is also generally leading within the field. 

\textbf{Task migration.} Lip Forensics \cite{LipForensics} transfers pre-trained models from lip-reading to Deepfake-generated facial video detection, while Shi et al. \cite{r10} and Kong et al. \cite{r11} propose models based on identifying real face distributions and Vision Transformer, respectively. This kind of approach leverages methods already existing in other forensic or visual tasks and adapts them for use in detecting Deepfake videos. The original pre-trained models undergo pre-training on large-scale datasets specific to other tasks, then fine-tuning on dedicated datasets for Deepfake detection. Compared to methods that directly train on Deepfake detection datasets, these models exhibit better generalization and robustness. 

\textbf{Generate data-driven.} Shiohara et al. \cite{r12} engage in facial forgery using source and target images, directing detection models to focus more on the forgery, and Chen et al. \cite{r13} aim to enrich the ``diversity'' of forgeries, thereby enhancing the ``sensitivity'' to forgeries. One of the most effective ways to enhance the performance of detection models is to provide training sets of equally high quality to assist in model training. However, in contrast to the high-level development of forgery techniques, the quality of existing datasets varies, and there is a lack of high-quality data. This situation has led to a dilemma in Deepfake detection, characterized by asymmetric adversarial challenges.

However, the aforementioned methods experience a decline in performance when detecting videos in real-world forgery scenarios. Currently, there are also a few works targeting real-world scenarios, such as compression. F3-Net \cite{F3-Net} enhances the frequency domain of frames affected by the Deepfake forgery process using frequency-aware decomposition and local frequency statistics. However, this method achieves good detection results only for videos with a specific compression rate. FT-two-stream \cite{FT-two-stream} proposes a dual-stream network for detecting compressed videos, but it obtains the compression dataset through hard coding and does not utilize compressed videos from real-world scenarios. To address this issue, Marcon et al. \cite{r15} download videos from social networks and fine-tune the network to detect compressed videos. Nevertheless, this method has a strong data dependency, and the influence of unknown platform types on model performance is significant. In response to this problem, Le et al. \cite{r16} apply frequency domain learning and optimal transport theory to knowledge distillation, thereby enhancing the model’s detection performance for low-quality videos. FAMM \cite{FAMM} characterizes facial muscle movements from a geometric perspective to detect compressed videos. However, all these methods focus only on FaceForensics++ (FF++) \cite{FF++}. The visual quality of the videos in this dataset is generally poor. It is not comparable to the highly realistic faked data currently circulating in social networks. In view of this, further research is conducted in this paper.
\section{THEORETICAL ANALYSES}
\subsection{Analysis of compressed Deepfake videos}
With the widespread dissemination of video content on social networks, a massive amount of unstructured video data has emerged. In this context, to ensure efficient transmission and storage of videos, social platforms commonly employ video compression technology. The primary purpose of video compression is to reduce the size of video files by removing redundant information (including spatial, temporal, and visual redundancies), thereby lowering transmission and storage costs. However, the compression process often introduces some noise and distortion. These not only result in compression artifacts overlapping with tampering artifacts in Deepfake videos but also hinder deep learning models from fully exploiting lost detail information. Additionally, the extra noise interferes with model training, increasing the difficulty of detecting Deepfake videos.

This study randomly selected a total of 300 uncompressed videos, hard-coded compressed videos, and social network compressed videos from FaceForensics++. Six metrics, namely Peak Signal-to-Noise Ratio (PSNR), Structural Similarity Index (SSIM), Universal Image Quality Index (UQI), Image Enhancement Factor (IEF), Visual Information Fidelity (VIF), and Relative Edge Change Ratio (RECO), were used to analyze the quality of videos before and after compression. PSNR is one of the commonly used metrics to measure image quality. It evaluates the distortion of images by calculating the mean squared error between the original image and the corrupted image, and then converts the result into a logarithmic ratio in decibels. Higher PSNR values indicate less image quality loss. SSIM is a metric used to measure the similarity between two images. It considers aspects such as brightness, contrast, and structure similarity, which are closely related to the human perception of image quality. The SSIM values range from -1 to 1, with values closer to 1 indicating higher similarity between the images. UQI is a universal image quality evaluation metric that comprehensively considers contrast, brightness, and structural information of images. Similar to SSIM, UQI evaluates image quality by comparing the similarity between the original and distorted images. IEF is used to evaluate the effectiveness of image enhancement algorithms. It compares the quality difference between the enhanced image and the original image to reflect the effectiveness of the enhancement algorithm. VIF is a metric used to evaluate video quality by comparing the structural similarity between the original video and the distorted video. RECO is also used to evaluate the impact of video distortion on edge information. It calculates the degree of video distortion by comparing the edge information between the original video and the distorted video. As shown in the Table \ref{table:1}, the PSNR values of both compressed videos are below the common threshold of 40 dB, indicating significant visual quality loss and detail information missing after compression. SSIM, UQI, and IEF values below 1 indicate the introduction of noise in the compressed videos. VIF values are 0.48 and 0.80, respectively, indicating the structural impact on compressed videos. RECO values are 0.85 and 1.00, respectively, indicating edge information loss in the compressed videos.

 \begin{table}[t!]
	\centering
	\caption{Quality assessment of hard-coded video (Faceforensucs++ LQ) and compressed video (Faceforensucs++ Social) in real scenes using PSNR, SSIM, UQI, IEF, VIF, RECO.
	}
	\label{table:1}
	\setlength{\tabcolsep}{3.5pt}
	\begin{tabular}{ccccccc}
		\toprule
		Datasets   & PSNR  & SSIM  & UQI & IEF  & VIF & RECO   \\
		\midrule
		Raw vs LQ & 32.06  & 0.84 & 0.99 & 0.91 & 0.48 & 0.85 \\
		Raw vs Social & 31.06 & 0.75 & 0.82 & 0.93 & 0.80 & 1.00 \\
		\bottomrule
	\end{tabular}
\end{table}

\begin{figure}[]
	\setlength{\abovecaptionskip}{-0.05cm}
	\begin{minipage}[b]{0.87\linewidth}
		\centering
		\includegraphics[width=0.87\linewidth]{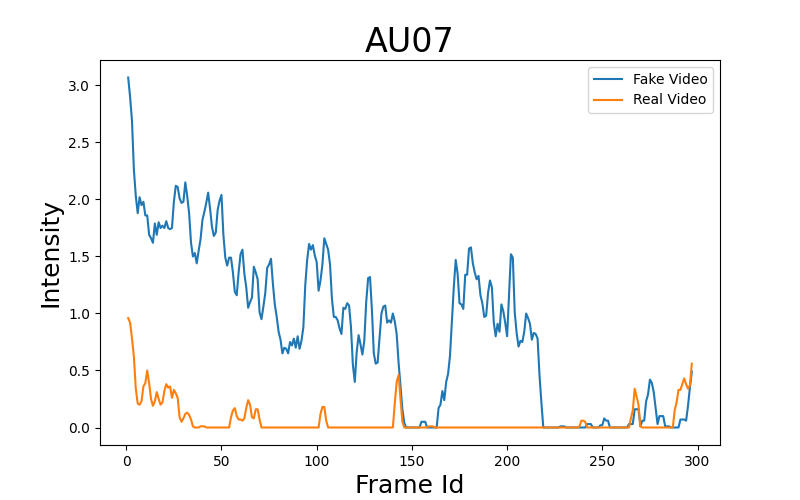}
		\vspace{0.04cm} 
	\end{minipage}
	\caption{The statistical intensity of Action Units (AU07) is tracked in 100 real videos and 100 fake videos, which describes upward eyelid movement. There is a significant contrast of the intensity between the real and fake videos.}
	\label{fig:AU}
\end{figure}

\begin{figure*}
	\centering
	\includegraphics[width=0.8\linewidth]{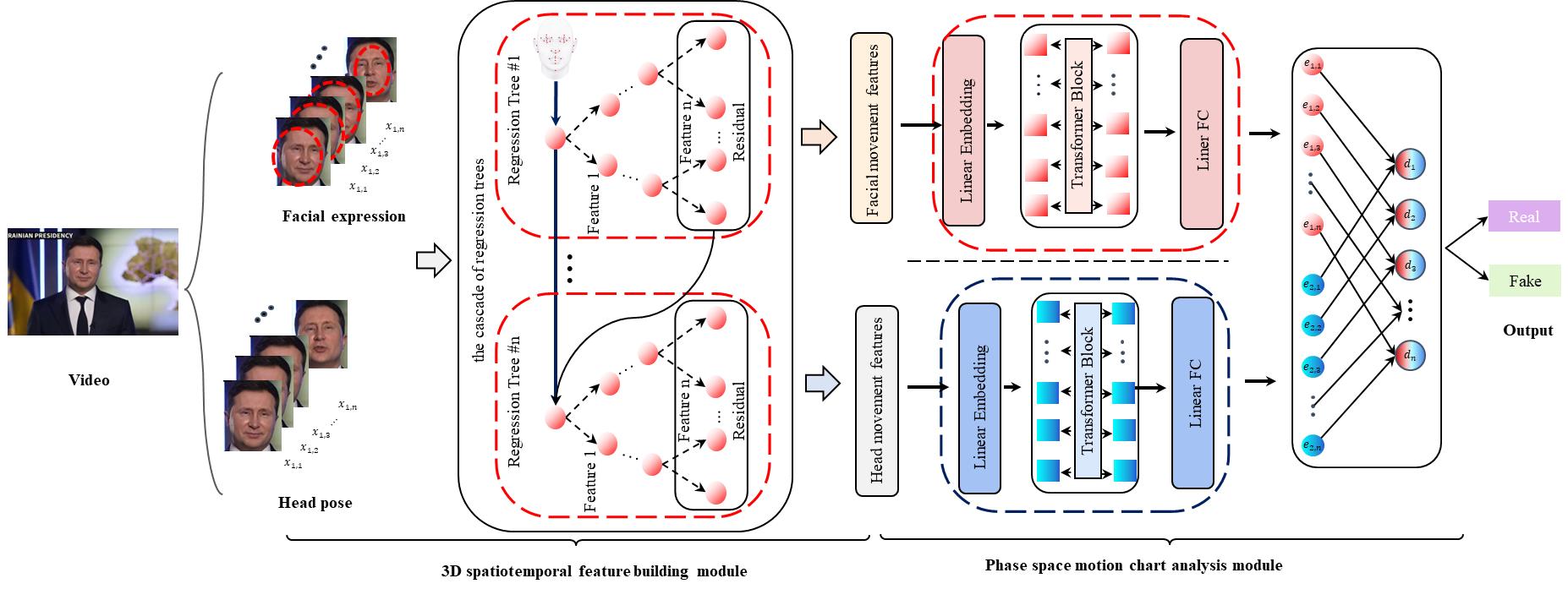}
	\caption{The overview of compressed Deepfake video detection based on 3D spatiotemporal. It consists of the 3D spatiotemporal feature construction module and the phase space motion map analysis module.}
	\label{fig:overview}
\end{figure*} 

\subsection{Analysis of facial motion}
In forged videos, there often exist unnatural facial movements, which may stem from multiple factors. Firstly, the forging process may lead to discontinuity or deformation of facial actions. In genuine videos, facial movements are typically continuous and natural, but in forged videos, due to potential abrupt changes, discontinuities, missing or distorted frames in each frame's synthesis, facial movements often appear discontinuous and lack smoothness. This lack of continuity is particularly evident in subtle facial movements such as eyelid movements. Secondly, the forging process results in a lack of coordination between facial movements and head postures. In genuine videos, facial movements often change with variations in head posture, reflecting the coordination between facial muscles and head movements. However, in forged videos, the limitations of synthesis algorithms may prevent accurate simulation of real head posture changes, or synchronization issues between facial and head movements may lead to a lack of coordination, resulting in unnatural facial expressions in the video. We utilized the open-source tool OpenFace 2.2.0 to track and locate facial features in the videos and analyze their temporal changes. We selected in 100 real videos and 100 fake videos from the FaceForensics++ (LQ) dataset for our study. Facial Action Coding System (FACS) is a system to taxonomize human facial movements by their appearance on the face. Movements of individual facial muscles are encoded by FACS from slight different instant changes in facial appearance. Using FACS it is possible to code nearly any anatomically possible facial expression, deconstructing it into the specific Action Units (AU) that produced the expression. It is a common standard to objectively describe facial expressions. In FACS, we focused on action units (AU07), which describes upward eyelid movement. Additionally, we examined head postures represented by Euler angles as 
\begin{equation}
   R = R_x \cdot R_y \cdot R_z
\end{equation}
where $R_x$, $R_y$, $R_z$ represent pitch, yaw, and roll, respectively. As depicted in Fig. ~\ref{fig:AU}, we observed relatively smooth temporal changes in genuine videos, while forged videos exhibited significant fluctuations and variations, highlighting their pronounced differences.
 
\section{METHODOLOGY}
\subsection{Overview}
Fig. ~\ref{fig:overview} illustrates the proposed Deepfake video detection framework based on 3D spatiotemporal trajectories. The method consists of two modules: 3D spatiotemporal feature construction module and phase space motion trajectory analysis module. In the 3D spatiotemporal feature construction module, a robust 3D model is employed for facial landmark localization and tracking, concurrently tracking head movements decoupled from facial expressions. Subsequently, through the spatial dynamics and temporal combination of facial action units (AU), spatiotemporal motion features are constructed. In the phase space motion trajectory analysis module, we process motion features using time-delayed embedding techniques to reconstruct phase space trajectories. Subsequently, we use these reconstructed trajectory data to train a lightweight Transformer architecture for exploring spatiotemporal patterns of facial features. Finally, we apply Dempster-Shafer evidence theory to fuse the model results. The core idea of this approach hinges on employing a resilient 3D landmark localization and tracking for crafting spatiotemporal motion features. In addition, this approach decouples head movements from facial expressions, covering a wide range of head movements, to evaluate facial muscle movements in a more nuanced way. These enables effective resilience against the influence of compressed videos on model detection performance.

\subsection{3D spatiotemporal feature building module}
The module utilizes a robust 3D model to localize and track facial landmarks and head movements, where head movements are decoupled from facial movements. Then the module selects facial points and head poses to construct phase space motion trajectories. These trajectories are used to characterize facial muscle movements in the temporal and spatial domains. Specifically, it first uses a 2D facial alignment algorithm to automatically locate 68 landmarks for each frame of the facial video. Secondly, the 3D facial model is used to estimate depth information from 2D frames, thereby enabling 3D landmark tracking. Finally, spatiotemporal domain motion features are constructed through the spatial dynamics and temporal combination of facial action units (AU). 

\subsubsection{Landmark localization and tracking}

Landmark localization, a cascade of trained regressors is employed to achieve accurate positioning of facial landmarks in each video frame. The gradient tree boosting algorithm is utilized to train each regressor, employing an accumulated square error loss. The training dataset comprises pairs $(I_i, S_i)$, where each $I_i$ represents a facial image, and $S_i$ denotes its corresponding shape vector. The initial shape estimation $S_i^{(0)}$ for each facial image is set to the mean shape of the training dataset: $S_i^{(0)} = \text{mean}(\{S_1, S_2, \ldots, S_n\})$. In each regression tree, the regression function $r_t$ is learned using the gradient tree boosting algorithm, and the estimate for each shape is updated as follows:
\begin{equation}
    S_i^{(t+1)} = S_i^{(t)} + r_t(I_i, S_i^{(t)})
\end{equation}
Notably, the initial shape selection at each level involves the use of Histograms of Oriented Gradients (HOG) features for centering and scaling, ensuring comparability across all frames.

Landmark tracking, the facial landmark tracking algorithm achieves decoupling of facial expressions and head poses, eliminating interference from head movements such as translation, scaling (approaching or moving away from the camera), and rotation (rolling, yawing, pitching) on facial expression landmarks. First, depth information for each facial pixel is extracted from 2D video frames using a 3D morphable face model (3DMM). This model incorporates Principal Component Analysis (PCA) for facial shapes, allowing the reconstruction of a 3D face from a single 2D image. The PCA model comprises principal components $V=[\upsilon_1, \upsilon_2, \ldots, \upsilon_n]$, the mean value of all facial meshes $\bar{\nu}$, and their standard deviation $\sigma_n$. The shape of a novel face is then generated as follows:
\begin{equation}
    S_i = \bar{\nu} + \sum_{n=1}^{N} \alpha_n \sigma_n \upsilon_n
\end{equation}
where $N$ is the number of principal components, and $\alpha_n$ represents $S_i$ in the coordinates of the PCA shape space. The 3D face shapes are reconstructed by fitting 68 detected landmarks to a PCA shape model. In the 3D model fitting stage, we employ the gold standard algorithm to achieve a least squares approximation of the affine camera matrix given a set of 2D-3D point pairs. By applying a 3D geometric transformation matrix, we are able to convert each frame into a frontal face, enabling tracking and comparison of facial landmark movements throughout the entire video. Simultaneously, we introduce geometric constraints for facial landmarks to address issues with the low accuracy of facial landmark detection algorithms in frames with significant head rotation angles or insufficient lighting.

\begin{figure}
	\centering
	\includegraphics[width=0.89\linewidth]{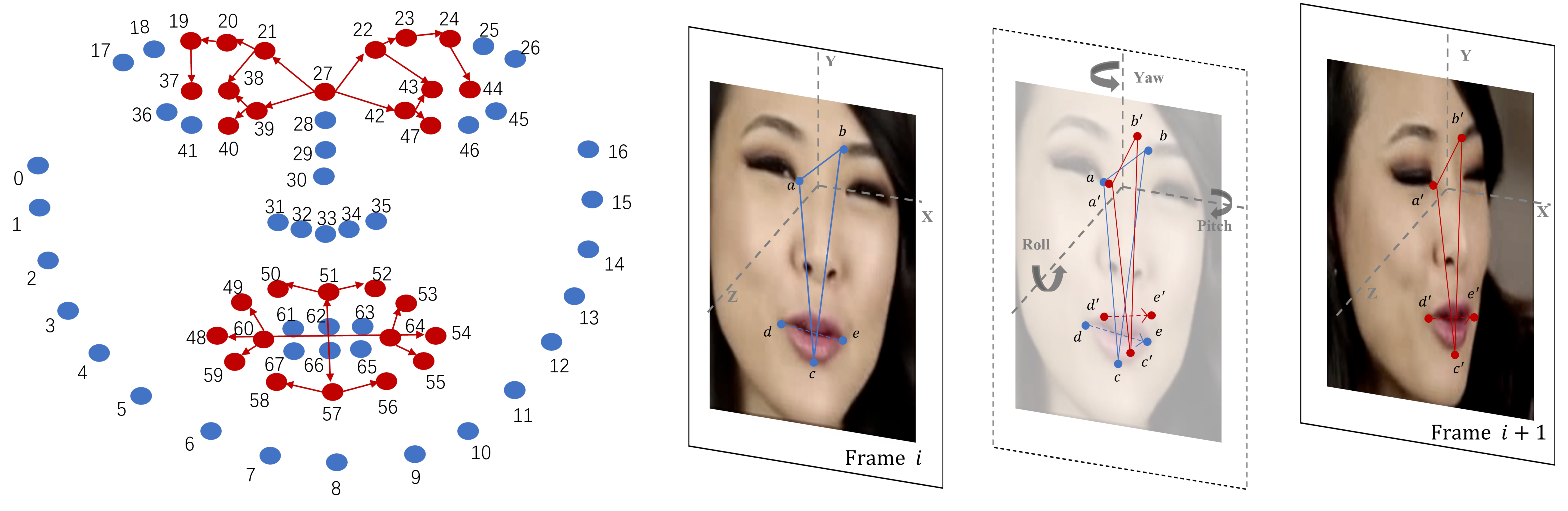}
	\caption{Construct spatial and temporal features based on 3D model. Left: Distance features and angular features of the eyebrow, eye, and mouth regions. Right: Rigid displacement and rotation angle features of the head in 3D space. }
	\label{fig:enter-label}
\end{figure}

\subsubsection{Construct spatial and temporal features based on 3D model}
By analyzing the spatial and temporal sequences of Facial Action Units (AU), facial movements can be quantified more reliably and specifically, while also helping to reduce random noise in landmark positioning. We select facial landmarks related to facial expressions and head poses to construct the following features, with a focus on the movement patterns of blinking, eyebrow raising, eye movement, lip closure, and head poses. (1) Eyebrow Region: Vertical positions of the left and right eyebrows, angle difference within the eyebrows. (2) Eye Region: Differences in eye corners, horizontal and vertical distances between eye corners. (3) Mouth Region: Horizontal distance between mouth corners, vertical distance of the lips, average vertical position between both mouth corners. (4) Head Displacement: Rigid displacements of the head in the X, Y, and Z directions. (5) Head Rotation: Rigid rotations of the head in the roll, pitch, and yaw directions. Construct spatial and temporal features based on 3D model as shown in Fig. ~\ref{fig:enter-label}.

The method proposed in this paper presents several advantages over other approaches in constructing facial movement features. Firstly, it achieves the decoupling of head movements from facial expressions. This decoupling allows subsequent deep learning models to more effectively capture the characteristics of either head movements or facial expressions, enabling them to focus on specific learning tasks without being influenced by other factors. This not only reduces the overall complexity of the task and improves accuracy but also makes it difficult for creators of forged videos to evade analysis of head movements even if they attempt to modify facial expressions to deceive detection models. Secondly, traditional AU detection algorithms typically only apply to frontal views of the face, while the 3D model tracking method proposed in this paper can directly and continuously extract facial and head movement information from video data. It considers cases with significant head rotation angles and ensures continuous measurement of landmarks. This provides a foundation for a more comprehensive representation of AUs and their intensity, rather than just a few discrete values. Consequently, it effectively mitigates the impact of compressed videos on model detection performance.

\subsection{Phase space motion chart analysis module}
This method first processes the constructed features by introducing temporal delay embedding to reconstruct phase space trajectories. Secondly, a recursive graph (RP) is created to capture relationships between features. Finally, a lightweight Transformer architecture is employed to explore the differences in feature distributions between real and fake videos in both temporal and spatial domains. Existing detection methods mainly focus on temporal differences between adjacent frames, while our approach concentrates on modeling facial variation patterns between the first frame and each subsequent frame within a continuous time segment, emphasizing holistic and global video feature analysis.

The recursive graph (RP) is a visualization method that uses a binary square matrix to represent temporal dependencies between all states in time series data. Assuming that at times \(i\) and \(j\) the states of system  \(X\) are represented by \(X_i\) and \(X_j\), respectively, recurrence can be recorded through a binary function as follows:
\begin{equation}
    R_{i,j}^X = \Theta(\epsilon_X - \|\mathbf{X}_i - \mathbf{X}_j\|_1), \quad \mathbf{X}_i \in \mathbb{R}^m, \quad i, j = 1, \ldots, N 
\end{equation}
where \(\Theta\) is a Heaviside function. For two time points \(i\) and \(j\) in the time series, if their similarity exceeds a predefined threshold \(\epsilon_X\), a point will be displayed at the corresponding position in the recursive graph (i.e., \(R_{i,j}^X = 1\)). By repeating this process over the entire time series, a matrix is created where each element represents the similarity between corresponding time points.

First, extract the feature motion trajectories between the first frame and subsequent frames of the video to construct a recursive graph. Secondly, convert the recursive matrix into the adjacency matrix of the network, representing the spatiotemporal neighborhood relationships between system states in the entire time series. Finally, to fully exploit spatiotemporal feature differences in real and fake videos, we design a lightweight Transformer classification model. The initial part of the model includes a linear layer to embed the input data into the hidden representation space. This allows the model to learn useful representations of the data without the need for manual feature engineering. Secondly, with the hidden dimension set to 128 and the number of encoder layers and attention heads to 2 each, a more lightweight Transformer model has been achieved. This model is suitable for practical deployment, particularly in scenarios with high task efficiency requirements and limited resources. Then, average pooling is employed to generate the final classification probability label instead of sequence-to-sequence output, thereby reducing the computational complexity of the model. Subsequently, calculate the loss between the predicted label and the actual label, and update the network parameters to complete the training. Finally, employ the Dempster-Shafer evidence theory to fuse the model results. In the testing phase, the algorithm receives the video to be tested as input and outputs the predicted label of the video. If the algorithm outputs 0, the video is predicted to be true; if the output is 1, the prediction is false.
\section{EXPERIMENTS}

\textbf{Datasets.} We conducted experiments on FaceForensics++ (FF++),  DFDC \cite{DFDC} and Celeb-DF \cite{r21} datasets. Videos from FF++ and Celeb-DF are compressed into two versions: medium compression (HQ) and high compression (LQ), using the H.264 codec with constant rate quantization parameters of 23 and 40 respectively. DFDC is the largest publicly available Deepfake detection dataset, which was released by Facebook in 2020.

\begin{table}[t]
	\renewcommand\arraystretch{1}
	\caption{Ablation study of the detection ACC (\%) on facial expressions (FE) and head pose (HP). }
	\label{table:2}
	\centering
	\setlength{\tabcolsep}{13pt}
	\begin{tabular}{cccc}
		\toprule[1.0pt]
		Datasets & w/o FE  & w/o HP & Ours\\
		\midrule[0.8pt]
		FF++HQ & 94.10 & 92.59 & \textbf{98.48} \\
		FF++LQ & 91.32 & 92.97 & \textbf{97.47} \\
		Celeb-DF-HQ & 94.92 & 96.45 & \textbf{99.37} \\
		Celeb-DF-LQ & 96.67 & 96.62 & \textbf{98.28} \\
		\bottomrule[1.0pt]
	\end{tabular}
\end{table}

\textbf{Implementation detail.} Firstly, frame-level assessment of the input video is performed using OpenCV to determine the presence of faces in each frame. Subsequently, Dlib is employed to locate 68 facial landmarks, enabling frame segmentation and annotation within the facial region. A regression tree cascade framework is utilized, where at each cascade level, the estimated landmark points are refined by adding the residuals generated by the previous regression tree. Next, a 3D facial model is employed to estimate depth information and pose estimation from 2D frames, facilitating the tracking of 3D landmarks. Finally, the network uses the Adam optimization algorithm to update model parameters with an initial learning rate of 0.0001. Learning rate adjustments are made using a scheduler based on the total loss for the current epoch when the loss function no longer decreases during training. The batch size is set to 128, and the number of epochs is 30. Loss, accuracy (ACC), area under the curve (AUC) and receiver operating characteristic (ROC) curve are computed at each epoch, and model parameters are saved based on the best loss. In our experiments, a dataset split of 8:1:1 is adopted, meaning that 80\% of videos are used for model training, 10\% for validation, and 10\% for testing.
\begin{table}[t]
	\renewcommand\arraystretch{1}
	\caption{The comparison of the detection ACC (\%) and AUC (\%) with the state-of-the-art methods on FF++. The experimental setup strictly follows TALL-Swim \cite{tall} and adopts its experimental results.}
	\label{table:3}
	\centering
	\begin{tabular}{ccccc}
		\toprule[1pt]
		\multirow{2}{*}{Method} & \multicolumn{2}{c}{FF++HQ}  & \multicolumn{2}{c}{FF++LQ} \\
		\cmidrule(r){2-3} \cmidrule(r){4-5} 
		& ACC & AUC & ACC & AUC \\
		\cline{1-5}
		MesoNet \cite{mesonet} & 83.10 & - & 70.47 & -\\
		Xception \cite{Xception} & 95.73 & 96.30 & 86.86 & 89.3\\
		Face X-ray \cite{Facex-ray} & - & 87.35 & - & 61.60\\
		Two-branch \cite{Two-branch} & 96.43 & 98.70 & 86.34 & 86.59\\
		Add-Net \cite{Add-net} & 96.78 & 97.74 & 87.50 & 91.01\\
		F3-Net \cite{F3-Net} & 97.52 & 98.10 & 90.43 & 90.43\\
		FDFL \cite{FDFL} & 96.69 & 99.30 & 89.00 & 92.40\\
		Multi-Att \cite{multi} & 97.52 & 98.10 & 90.43 & 90.43\\
		FT-two-stream \cite{FT-two-stream} & 92.47  &  95.56 & 90.70  & 91.25 \\
		FInfer \cite{hu2022finfer} & 95.67  &  97.17 & 92.27  & 93.10 \\
		RECCE \cite{RECCE} & 97.06 & 99.32 & 91.03 & 95.02\\
		FAMM \cite{FAMM} & 96.75 &  97.98 &  94.67 & 96.98 \\
		TALL-Swim \cite{tall} & \textbf{98.65} & \textbf{99.87} & 92.82 & 94.57\\
		\textbf{Ours} & 98.48 & 98.82 & \textbf{97.47} & \textbf{97.98}\\
		\bottomrule[1pt]
	\end{tabular}
\end{table}
\begin{table}[t!]
	\renewcommand\arraystretch{1}
	\caption{The comparison of the detection ACC (\%) and AUC (\%) with the state-of-the-art methods on Celeb-DF and DFDC. The experimental setup strictly follows BRCNet \cite{BRCNet} and adopts its experimental results.}
	\label{table:4}
	\centering
	\begin{tabular}{ccccc}
		\toprule[1pt]
		\multirow{2}{*}{Method} & \multicolumn{2}{c}{Celeb-DF-V2-HQ}   & \multicolumn{2}{c}{DFDC} \\
		\cmidrule(r){2-3} \cmidrule(r){4-5} 
		& ACC & AUC  & ACC & AUC\\
		\cline{1-5}
		Two-stream \cite{Two-stream} & - & 53.8 & -  & 61.40 \\
		Xception \cite{Xception} & 97.9 & 99.73 & 79.35  & 80.92 \\
		Face X-ray \cite{Facex-ray} & - & - & 72.65  & 89.5\\
		Multi-Att \cite{multi} & 97.92 & 99.94 &76.81&90.32\\
		Add-Net \cite{Add-net} & 96.93 & 99.55  &  78.71& 89.85 \\
		F3-Net \cite{F3-Net} & 95.95 & 98.93  & 76.17 &88.39  \\
		FT-two-stream \cite{FT-two-stream} & 80.74  & 86.67  & 63.85   &  64.03 \\
		FInfer \cite{hu2022finfer} & 90.47  & 93.30  & 80.39   &  82.88 \\
		RECCE \cite{RECCE} & 98.59 & \textbf{99.94}  &   81.20&  91.33\\
		FAMM \cite{FAMM} & 82.54 & 85.33  &  80.10   &  84.25 \\
		BRCNet \cite{BRCNet} & 98.73 & 99.94 &81.65  &91.89  \\
		\textbf{Ours} & \textbf{99.37} & 99.56 &\textbf{93.29} &\textbf{93.51}\\ 
		\bottomrule[1pt]
	\end{tabular}
\end{table}
\begin{figure*}[t]
	\centering
	\subfloat[FF++HQ]{\includegraphics[width=0.5\columnwidth]{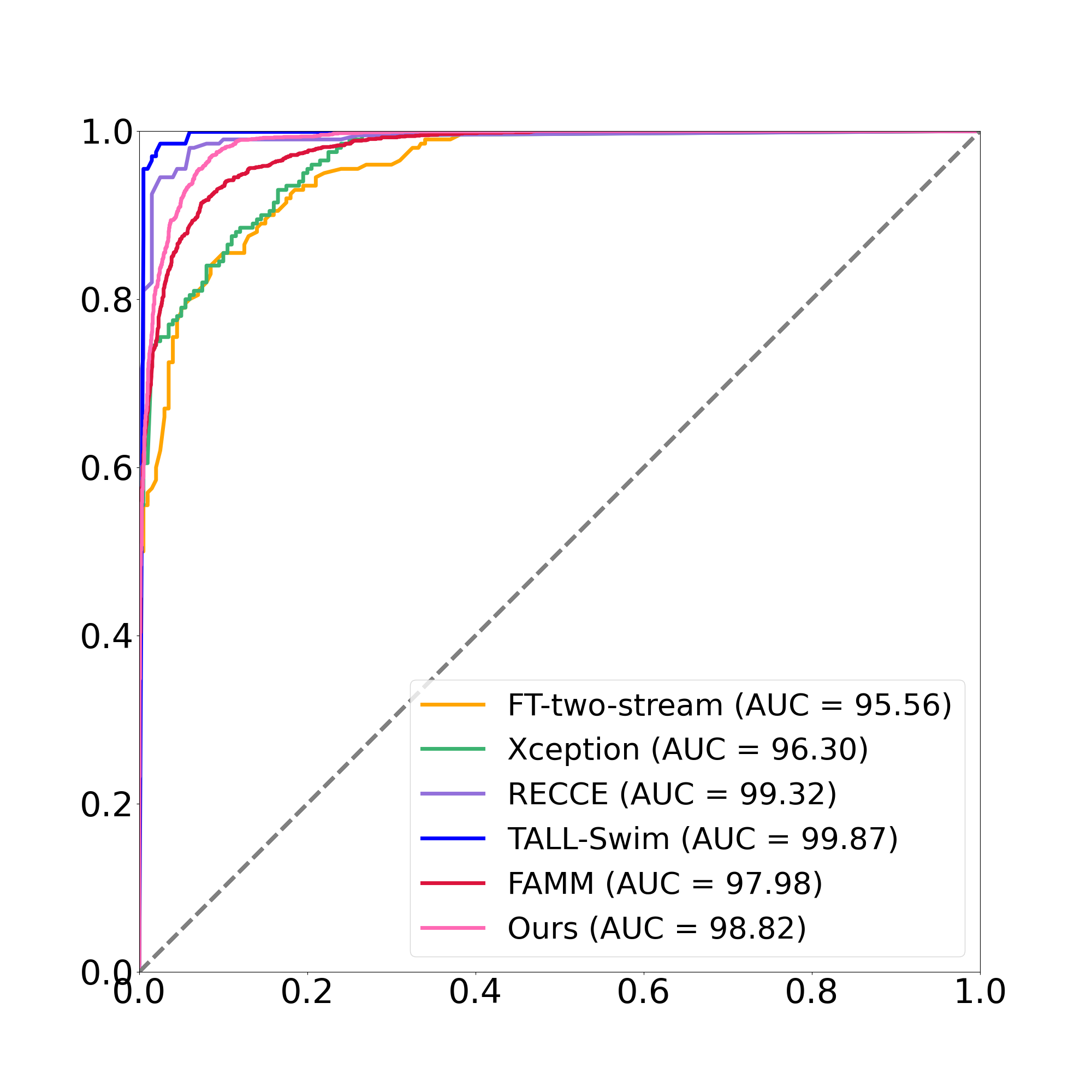}}
	\subfloat[FF++LQ]{\includegraphics[width=0.5\columnwidth]{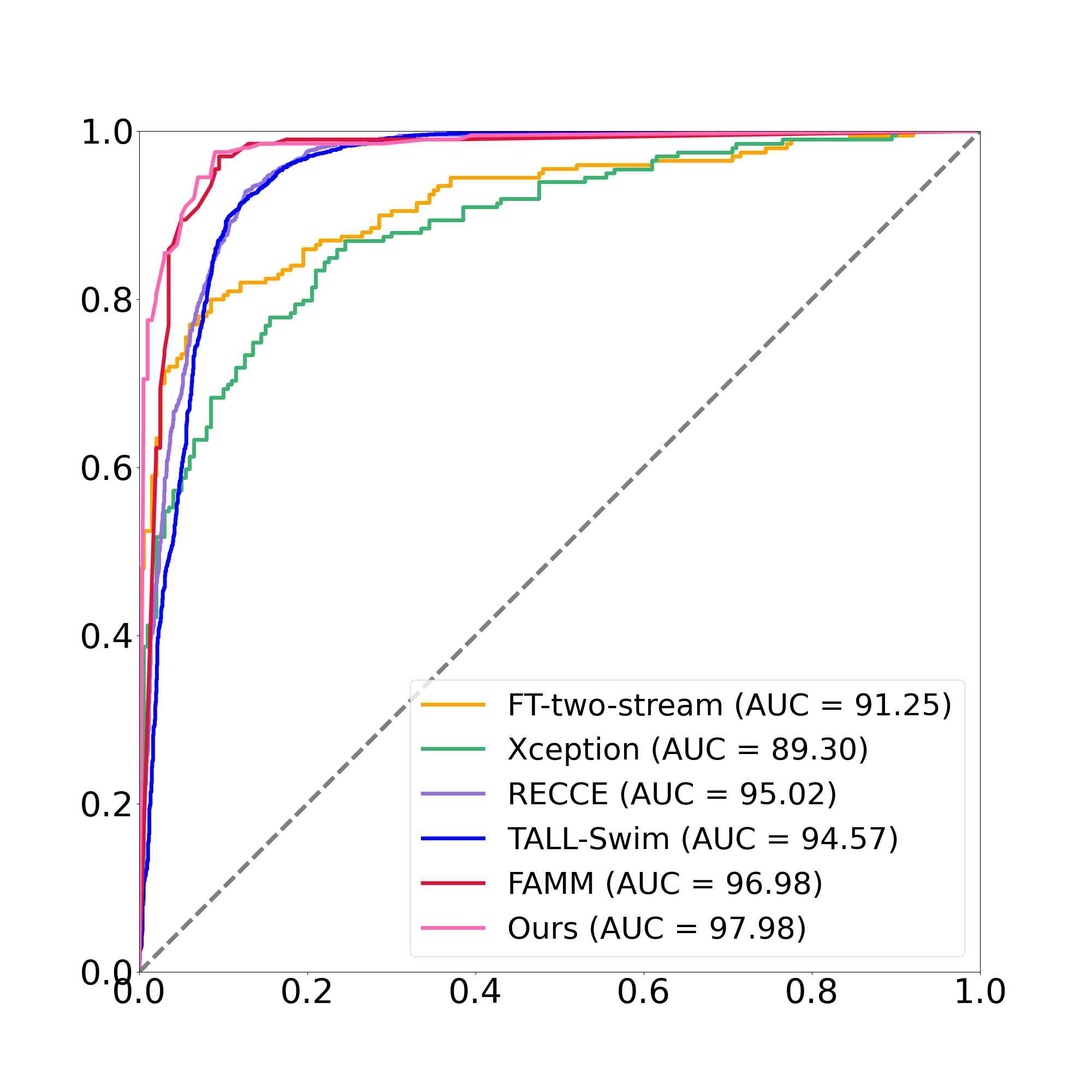}}
	\subfloat[Celeb-DF-V2-HQ]{\includegraphics[width=0.5\columnwidth]{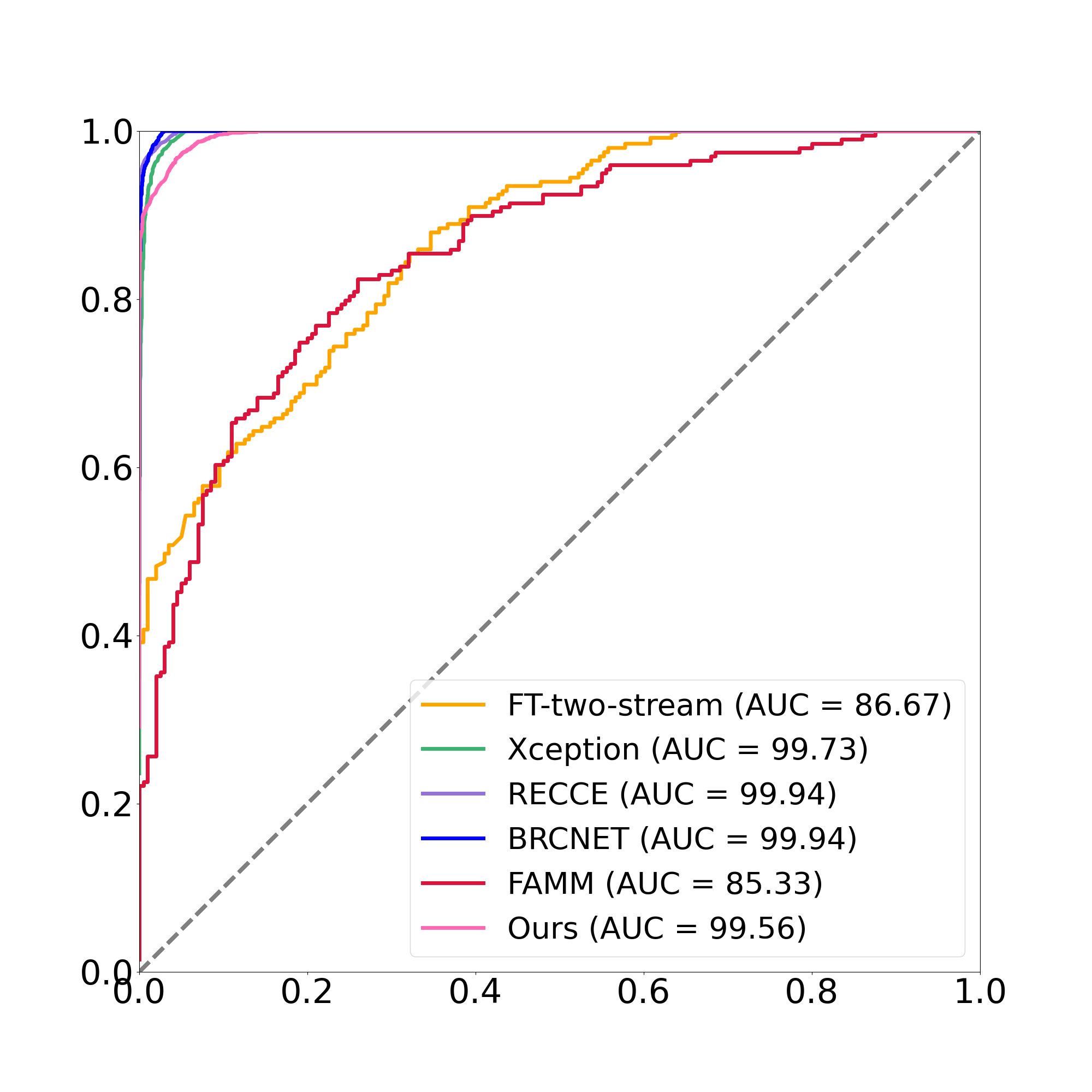}}
	\subfloat[DFDC]{\includegraphics[width=0.5\columnwidth]{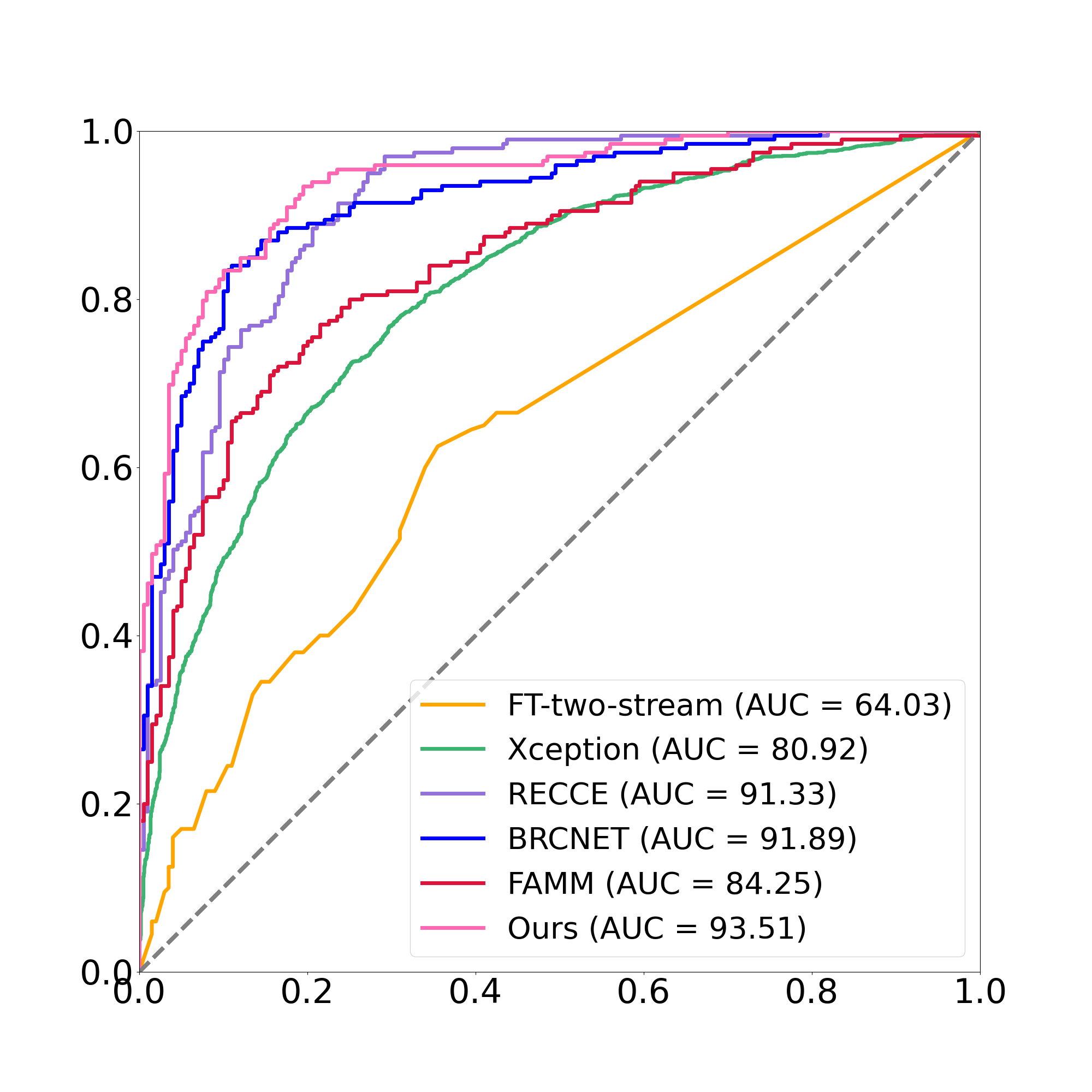}}
	\caption{ROC (receiver operating characteristic) curves for the state-of-the-art compressed Deepfake videos detection methods on different public datasets:
		(a) FF++HQ, (b) FF++LQ, (c) Celeb-DF-V2-HQ, (d) FF++LQ.}
	\label{fig:roc}
\end{figure*}

\textbf{Ablations.} This paper characterizes the relative movement patterns of facial landmarks in both temporal and spatial domains by integrating facial expressions and head pose. We conducted ablation studies to assess the combination of facial expressions and head pose. Specifically, we utilized only facial expressions or head pose to represent facial motion, and the results are presented separately in Table ~\ref{table:2}. Experimental results indicate that the detection performance achieved by combining facial expressions and head pose surpasses that of using only facial expressions or head pose. The reason for this improvement may be attributed to their combination, which provides an excellent feature for facial motion.

\textbf{Comparison with SOTA methods.} Table ~\ref{table:3} shows the ACC and AUC evaluation metrics of our method and existing methods on the FaceForensics++ dataset. We conducted comparative experiments on HQ and LQ. In the LQ dataset, our method is outperforming TALL-Swim \cite{tall} by 4.65\% in terms of the ACC and 3.41\% in terms of the AUC. And the performance on HQ is currently comparable to the existing methods. Table ~\ref{table:4} displays the ACC and AUC evaluation metrics of our method and existing methods on the Celeb-DF and DFDC datasets. Our method achieves an ACC value of 93.29 and an AUC value of 93.51 on DFDC, surpassing existing methods. And it demonstrates competitive detection performance on the Celeb-DF-V2-HQ. We also evaluate the overall detection performance using the ROC (receiver operating characteristic) curve, and the results are shown in Fig. ~\ref{fig:roc}. The abscissa values represent the FPR (False Positive Rate), and the ordinate values represent the TPR (True Positive Rate). Our curve is closer to the top left hand corner, representing our method is better than the state-of-the-art methods on the compressed DeepFake videos detection.

\begin{table}[t]
	\begin{center}
		\renewcommand{\tabcolsep}{2mm} 
		\caption{Training and testing time comparison with state-of-the-art methods.}
		\label{table:eff}
		\begin{tabular}{cccc}
			\toprule[1.0pt]
			Method          & Training Time (s) & Testing Time (s) \\
			\midrule[0.8pt]
			FWA \cite{FWA}             & 14300              & 197              \\
			MesoNet \cite{mesonet}         & 2007               & 70               \\
			Capsule \cite{Capsule}         & 36600              & 125              \\
			Re-net \cite{Re-net}          & 1695               & 6                \\
			FAMM \cite{FAMM} & 1033               & 15               \\
			Ours            & \textbf{451}       & \textbf{3.33}     \\
			\bottomrule[1.0pt]
		\end{tabular}
	\end{center}
\end{table}

\begin{figure*}[t]
	\centering
	\subfloat[Fake video]{\includegraphics[width=0.7\columnwidth]{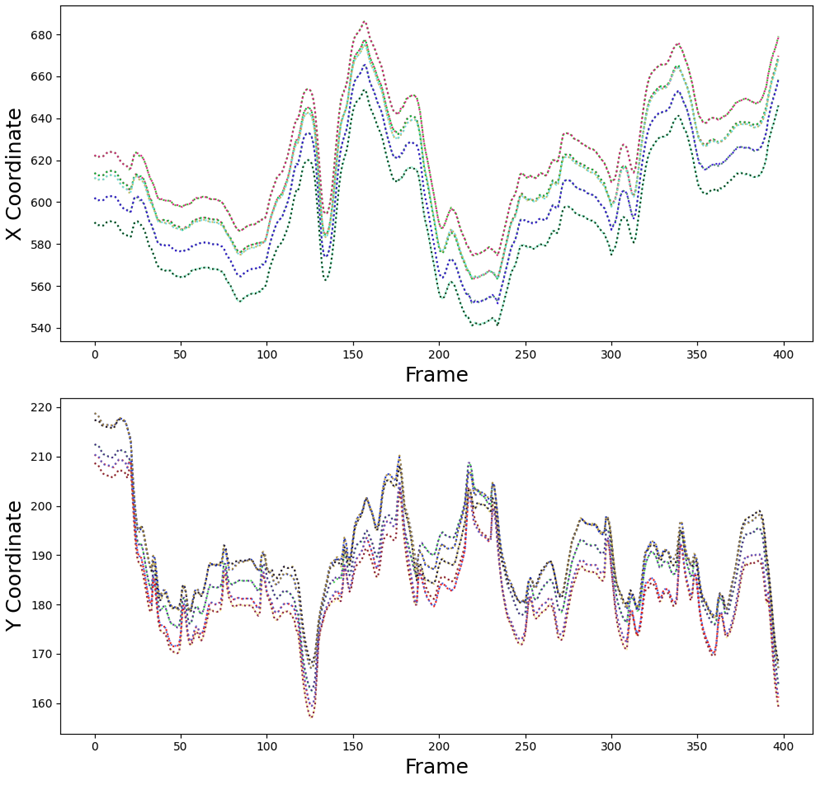}}
	\subfloat[Real video]{\includegraphics[width=0.7\columnwidth]{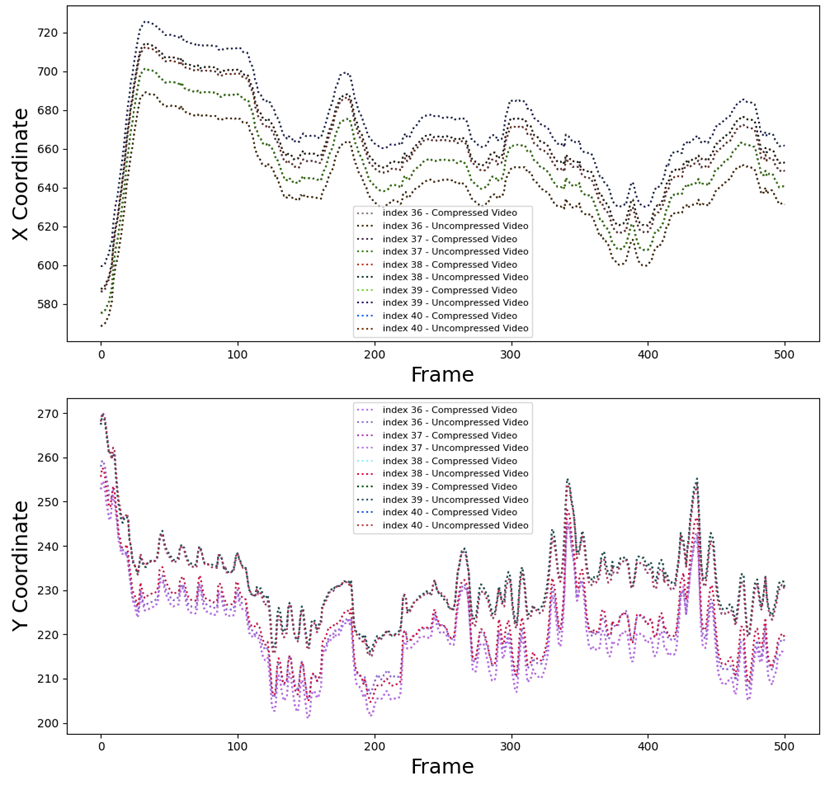}}
	\caption{The changes in landmarks before and after video compression are depicted. (a) illustrates the continuous changes in x and y coordinates in the Fake video, (b) illustrates the continuous changes in x and y coordinates in the Real video. The observed results indicate that the distributions of landmarks before and after compression almost completely overlap, suggesting that video compression does not alter the distribution of facial coordinate points.}
	\label{fig:com}
\end{figure*}
\begin{figure*}[t]
	\centering
	\subfloat[Fake video]{\includegraphics[width=0.7\columnwidth]{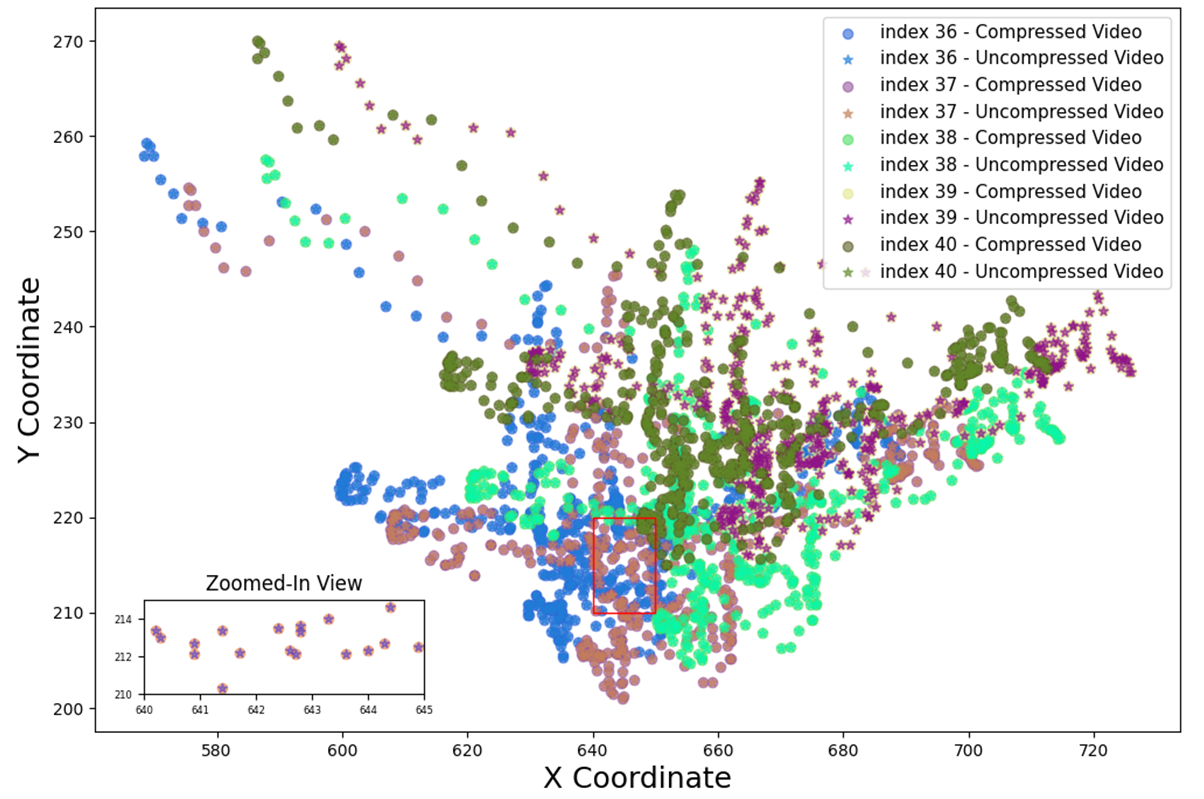}}
	\subfloat[Real video]{\includegraphics[width=0.7\columnwidth]{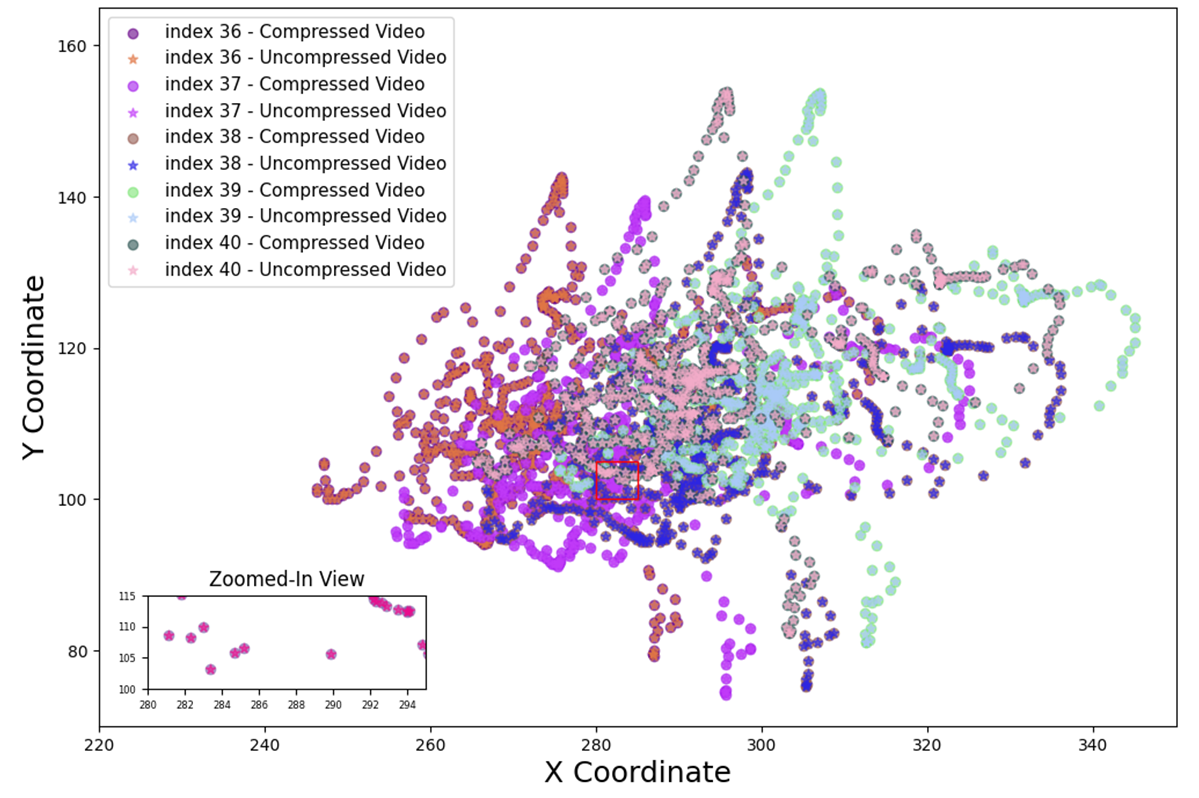}}
	\caption{The 2D visualization of video coordinate distribution. The circle `o' represents the coordinates of the compressed video and the asterisk `*' represents the coordinates of the uncompressed video.}
	\label{fig:san}
\end{figure*}

Compared to the existing baselines, our method achieves nearly optimal performance in video compression experiments. We attribute this to the following reasons. On the one hand, almost no landmark errors are introduced during the video compression process, which does not affect the designed features. However, facial landmark errors are introduced during video tampering, altering facial motion patterns. Our method directly employs a robust 3D model to locate and track facial and head landmarks in videos, and then constructs features combining 2D and 3D frames. This enhances the robustness of the model. To confirm the above points, we randomly select 100 real and 100 fake videos from FaceForensics++, and analyze the distribution of landmarks before and after compression. Specifically, we selected landmarks in the left eye region, where the landmarks are numbered 36 to 40. As shown in Fig. ~\ref{fig:com}, the changes in x-coordinates and y-coordinates of the real and fake videos before and after compression are basically the same, thus confirming that compression does not change the distribution of facial landmarks. In order to observe the changes in the distribution of coordinate points more clearly, we performed a 2D visualization of the coordinate points of the video. As shown in the Fig. ~\ref{fig:san}, it can be clearly seen that the circle `o' tightly wraps the asterisk `*', which intuitively illustrates that the coordinate point distribution of the real and fake videos before and after compression does not change, further confirming that compression does not change the distribution of facial landmarks. On the other hand, existing detection methods mostly rely on neural networks. The video compression introduces compression artifacts that coexist with tampering artifacts, which may mislead baseline learning. Therefore, our method performs better in detecting compressed videos.

\textbf{Detection efficiency.}
The time complexity of the algorithm is as follows. Let's assume the number of input videos is \(m\). In the landmark localization and tracking stage, as well as the feature extraction stage, the algorithm only requires traversing the number of input videos. The remaining loops are all constants. Therefore, the time complexity of the algorithm is \(O(m)\). Additionally, comparing the training and testing times of our model with state-of-the-art methods, the results are shown in the Table ~\ref{table:eff}. The training and testing times required by our method are minimal. In summary, our method demonstrates superior detection efficiency, making it conducive for the deployment of models in practical scenarios.

\section{CONCLUSIONS}
This paper pioneers the migration of 3D models into the task of Deepfake facial landmark localization and tracking, constructing more robust facial motion features. Additionally, through a sequential analysis approach based on phase space motion trajectories, it explores the overall and global features of Deepfake videos. Finally, extensive experiments demonstrate that our method achieves state-of-the-art performance on compressed videos and also performs well on uncompressed videos. At the same time, our method exhibits the highest detection efficiency that is more suitable for practical applications in real-world scenarios. Future work will focus on further improving the algorithm's robustness and deploying it for widespread use in real-world scenarios.

\bibliographystyle{ACM-Reference-Format}
\bibliography{sample-base}

\end{document}